\documentclass{article}
\usepackage{ijcai25}

\usepackage{times}
\usepackage{soul}
\usepackage{url}
\usepackage[hidelinks]{hyperref}
\usepackage[utf8]{inputenc}

\usepackage[small]{caption}
\usepackage{graphicx}
\usepackage{amsmath}
\usepackage{amsthm}
\usepackage{booktabs}
\usepackage{algorithm}
\usepackage{algorithmic}
\usepackage[switch]{lineno}
\usepackage{subcaption} 
\newcommand{\indep}{\mathrel{\perp\!\!\!\perp}}

\newtheorem{theorem}{Theorem}
\newtheorem{assumption}{Assumption}
\newtheorem{Definition}{Definition}

\title{Deconfounding Multi-Cause Latent Confounders: A Factor-Model Approach to Climate Model Bias Correction}

\author{
Wentao Gao$^{1}$
\and
Jiuyong Li$^{1}$
\and
Debo Cheng$^{1}$
\and
Lin Liu$^{1}$
\and
Jixue Liu$^{1}$
\and
Thuc Le$^{1}$
\and
Xiaojing Du$^{1}$
\and
Xiongren Chen$^{1}$
\and
Yun Chen$^{2}$
\and
Yanchang Zhao$^{3}$\\
\affiliations
$^1$University of South Australia, Adelaide, SA, Australia\\
$^2$CSIRO Environment, Canberra, Australia\\
$^3$CSIRO Data61, Canberra, Australia\\
\emails
\{wentao.gao\}@mymail.unisa.edu.au\\
}

\begin{document}

\maketitle

\begin{abstract}
Global Climate Models (GCMs) are crucial for predicting future climate changes by simulating the Earth systems. However, the GCM Outputs exhibit systematic biases due to model uncertainties, parameterization simplifications, and inadequate representation of complex climate phenomena. Traditional bias correction methods, which rely on historical observation data and statistical techniques, often neglect unobserved confounders, leading to biased results. This paper proposes a novel bias correction approach to utilize both GCM and observational data to learn a factor model that captures multi-cause latent confounders. Inspired by recent advances in causality based time series deconfounding, our method first constructs a factor model to learn latent confounders from historical data and then applies them to enhance the bias correction process using advanced time series forecasting models. The experimental results demonstrate significant improvements in the accuracy of precipitation outputs. By addressing unobserved confounders, our approach offers a robust and theoretically grounded solution for climate model bias correction.
\end{abstract}

\section{Introduction}

Global Climate Models (GCMs), such as those developed by the Coupled Model Intercomparison Project phase 6 (CMIP6) \cite{CMIP6}, are vital tools for predicting future climate changes. These models simulate the physical and chemical processes of the Earth systems—including the atmosphere, oceans, land, and ice—to provide detailed climate forecasts. Despite significant advancements in GCM, their output still exhibits systematic biases. These biases primarily come from uncertainties within the models, simplifications in parameterization processes, and inadequate representations of complex climate phenomena \cite{Mouatadid2023}. For instance, processes such as cloud formation, precipitation, radiation, and convection are often simplified into parameterization formulas in GCMs, which may not accurately reflect real-world conditions, thus leading to biases. Consequently, \cite{Lafferty2023} highlighted that bias corrections were particularly crucial for near-term precipitation projections, especially in cases where observational data are inconsistent with the GCM Output.

To enhance the reliability and accuracy of the GCM Outputs, numerous bias correction techniques, ranging from simple linear scaling to advanced quantile mapping, have been proposed to modify the GCM Outputs and align them more closely with actual observations \cite{Casanueva2020,Chen2013,Dowdy2020,Enayati2020,Feigenwinter2018,Lafon2013,Maraun2016,Mehrotra2018,Miao2016,Nahar2018,Piani2010,Smitha2018,Teutschbein2012,Wu2022}. These methods typically rely on historical observation data and statistical approaches to adjust climate model outputs and correct systematic errors \cite{Maraun_Widmann_2018}.

\begin{table*}[ht]
\centering
\renewcommand{\arraystretch}{1.5}
\begin{tabular}{|c|c|c|c|}
\hline
\textbf{Time period} & \textbf{History} & \textbf{Current} & \textbf{Future} \\
\hline
 & $\left[ t - h - w - 1, \ldots, t - w - 1 \right]$ & $\left[ t - w, \ldots, t \right]$ & $\left[ t + 1, \ldots, t + k + 1 \right]$ \\
\hline
\textbf{GCM(G)} & $\mathbf{X}^G_t (\text{humidity, pressure,} \ldots)$ & $\mathbf{A}^G_t (\text{humidity, pressure,} \ldots)$ & ${Y}^G_t (\text{precipitation})$ \\
\hline
\textbf{Observations(O)} & $\mathbf{X}^O_t (\text{humidity, pressure,} \ldots)$ & $\mathbf{A}^O_t (\text{humidity, pressure,} \ldots)$ & ${Y}^O_t (\text{precipitation})$ \\
\hline
\end{tabular}
\caption{A description of data variables for GCM and observations, and \(\mathbf{V} = \{\text{humidity}, \text{temperature}, \text{pressure}, \text{precipitation}, \ldots\}\) }
\label{table:gcm_observations}
\end{table*}

However, these traditional methods have been found to inflate simulated extremes, raising concerns about their use in climate change applications where extremes are significant, such as drought and flooding \cite{PastenZapata2020}. Another major limitation of most bias correct methods is their assumption that all relevant factors are known and observable, which is unrealistic in practical applications. Climate systems are complex and include many factors that may not be fully observed, such as microclimate effects, regional climate characteristics, and anthropogenic influences. These potential unobserved (confounding) factors dynamically impact time series forecast. However, since they cannot be fully observed, they are frequently overlooked by current bias correction methods. This overlook limits the potential of most existing methods in bias correction.

To deal with the challenge of extreme conditions, the study by \cite{nivron2024temporalstochasticbiascorrection} incorporates advanced time series forecasting models into the bias correction of extreme weather events, such as heatwaves, by considering the bias correction as a time-indexed regression model with stochastic output. This provides a new perspective: adapting time series forecasting models to bias correction can improve model performance. However, although advanced models have shown significant potential in time series forecasting \cite{zhou2021informer,wu2022autoformer,wu2023timesnet,wang2023timemixer,gao2024tsimultiviewrepresentationlearning}, particularly in climate forecasting \cite{Bi2023,Wu2023}, they typically overlook the existence of unobserved confounding factors, which leads to biased results.

In the field of causal inference with time series, recent studies have started addressing the challenge of unobserved confounders in predicting the potential outcomes of treatments in time series data. Instead of assuming unconfoundedness \cite{Pearl2000}, these studies operate under a weaker assumption that only multi-cause confounders exist \cite{bica2020time,Wang2019,li2024identification,cheng2023instrumentalvariableestimationcausal}. A variable is said to be a confounder if it is a common cause of both the treatment and the outcome \cite{hernan2020causal}. For example, in the Directed Acyclic Graph (DAG) shown in Figure \ref{fig:causal graph}(a), the confounder \(\mathbf{X}\) affects both the treatment \(\mathbf{A}\) and outcome \(Y\). In our bias correction problem, precipitation output is always treated as the outcome, and other related climate variables are treated as treatment variables. The most common method for predicting potential outcomes is to use all treatment variables. However, neglecting unobserved multi-cause confounders \(\mathbf{Z}\), such as large-scale atmospheric circulation patterns or oceanic processes, can lead to simultaneous effects on various climate variables, including precipitation. These multi-cause confounders make it possible to apply current research outcomes to bias correction in GCM.

Inspired by the work of \cite{bica2020time}, in this paper, we propose a deconfounding bias correction method. It is important to note that estimating hidden confounders in climate bias correction is much more complex. This increased complexity arises not only from the intricate nature of hidden confounders in climate science but also because these confounders need to be inferred from GCM and observation data.

Our contribution can be summarized as the following:

\begin{enumerate}
\item We use causality to identify and understand unobserved confounders, allowing us to obtain an unbiased outcome in the presence of multi-cause confounders. By identifying latent confounders through constructing a factor model over time, our approach can capture these unobserved factors, resulting in unbiased outcomes.
\item We develop a two-phase algorithm: Deconfounding and Correction. The Deconfounding Bias Correction (BC) factor model captures the confounders from both Global Climate Model (GCM) output data and observational data, and the correction model uses these confounders as additional information for the bias correction task. Inspired by the TSD (Time Series Deconfounder) \cite{bica2020time}, our method extends to GCM and observation data.
\end{enumerate}

\section{Problem Formulation}

As shown in Table \ref{table:gcm_observations}, let the random variable \(\mathbf{V}\) represent the set of climate variables, such as humidity, temperature, pressure, precipitation and others. Among these variables, \(Y\) denote the outcome of interest, for example, precipitation. In our problem setting, because we will use both historical and current data of these variables, to avoid confusion, \(\mathbf{X}_t\) represents the historical version of \(\mathbf{V}\setminus{Y}\). \(\mathbf{A}_t\) represents the current version of \(\mathbf{V}\setminus{Y}\). 

These variables are collected from two sources: Global climate models (GCMs) and observations. 
Moreover, we use \(\mathbf{X_t^G}\) and $\mathbf{A_t^G}$ to represent historical and current variables of the GCM data respectively, while \(\mathbf{X_t^O}\) and \(\mathbf{A_t^O}\) for historical and current variables. The outcome from GCMs and observations are denoted as \(Y_t^G\) and \(Y_t^O\) respectively.

The data for a location, also known as the location trajectory, consist of realizations of the previously described random variables \(\{\mathbf{x_t^G}, \mathbf{a_t^G}, y_t^G, \mathbf{x_t^O}, \mathbf{a_t^O}, y_t^O\}\). Let \( y(\mathbf{\bar{a}_t}) \) represent the potential outcome (precipitation), which could be either factual or counterfactual, for each possible treatment course \(\mathbf{\bar{a}_t}\), where \(\mathbf{\bar{a}_t} = (\mathbf{a_1},...,\mathbf{a_t})\). Consequently, we have \( y^G(\mathbf{\bar{a}_t}) \) and \( y^O(\mathbf{\bar{a}_t}) \).
The concept of potential outcomes allows us to consider what the outcome \(y\) would be under different treatment scenarios, which is essential for causal inference \cite{Rubin1974}.

\begin{figure}[h]
    \centering
    \includegraphics[width=1.0\linewidth]{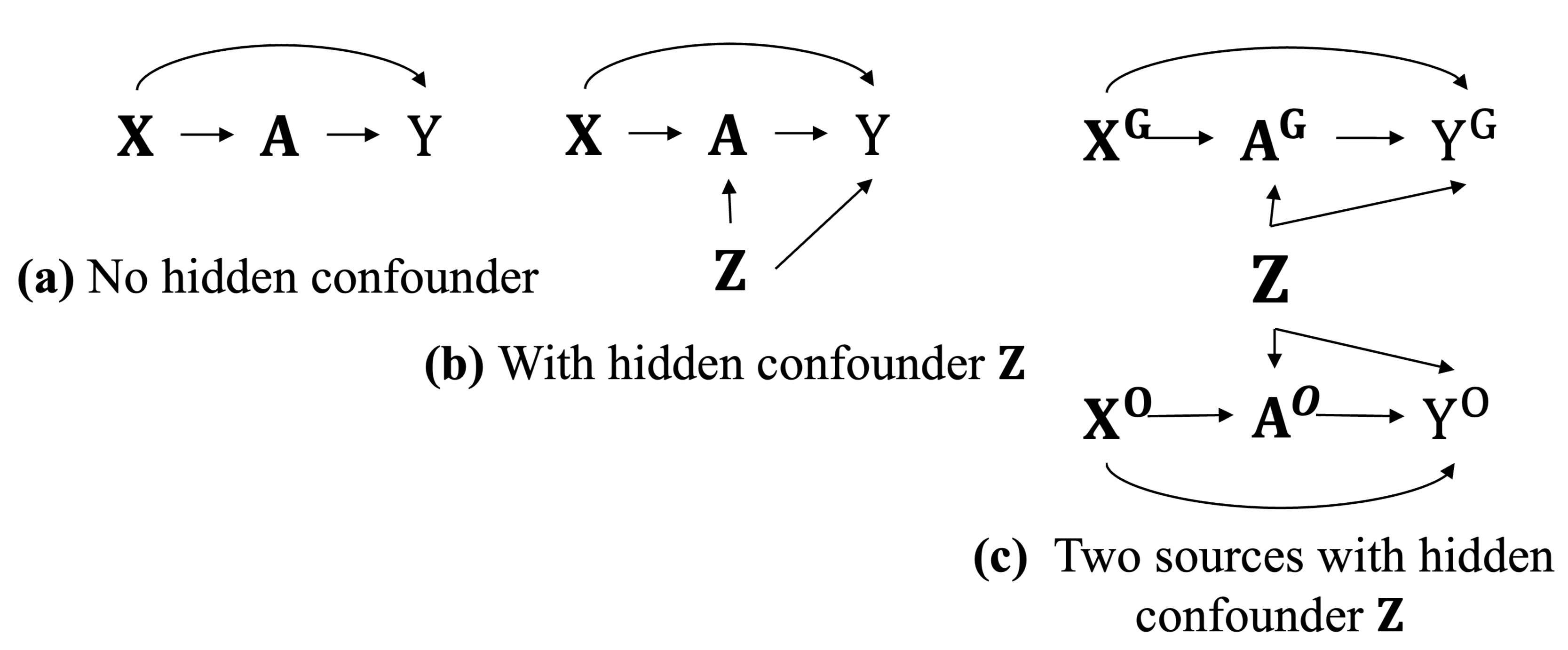}
    \caption{Summary causal graphs under three conditions: (a) No hidden confounder, (b) With hidden confounder $\mathbf{Z}$, (c) Two sources with hidden confounder $\mathbf{Z}$}
    \label{fig:causal graph}
\end{figure}

To correct future \(k\) steps precipitation bias \(\Delta{Y} = Y^O - Y^G\) which is the difference between observational precipitation \(y^O\) and GCM output precipitation \(y^G\), a common way is to use all data to obtain a regression model \cite{nivron2024temporalstochasticbiascorrection} which can be represented by Figure \ref{fig:causal graph}(a). However, the existence of hidden confounders (e.g. unobserved atmospheric and oceanic circulation) will result in biased result.  To address the issue of hidden confounders in time series data, \cite{bica2020time} have developed a method, assuming the presence of multi-cause hidden confounders \(\mathbf{Z}\), as shown in the Figure \ref{fig:causal graph}(b).

However, no attempt has been made to adapt their work to bias correction, including bias correction for climate models. In this paper, we extend the work for the bias correction problem as shown in the Figure \ref{fig:causal graph}(c) which makes the most of information we have. We assume that hidden variables \(\mathbf{Z}\) affect both GCM and observations. Learning \(\mathbf{Z}\), we would estimate the potential outcome in observations for each location conditional on the location history of covariates$\bar{\mathbf{X}}_t^O = (\mathbf{X}_1^O, \ldots, \mathbf{X}_t^O) \in \mathcal{X}_t^O$, treatments $\bar{\mathbf{A}}_t^O = (\mathbf{A}_1^O, \ldots, \mathbf{A}_t^O) \in \mathcal{A}_t^O$ and confounders $\bar{\mathbf{Z}}_t = (\mathbf{Z}_1, \ldots, \mathbf{Z}_t) \in \mathcal{Z}_t$:
\begin{equation}
    \mathbf{E}\left[ \mathbf{Y}^O(\mathbf{\bar{a}}^O_{\geq t}) \mid \mathbf{\bar{A}}^O_{t-1}, \mathbf{\bar{X}}^O_{t}, \mathbf{\bar{Z}}_{t}\right] 
\end{equation}

\section{Proposed Method}
The existence of confounders can result in the obtained association among \(\mathbf{X, A}, Y\) not accurately representing the true relationships, potentially leading to biased results. To solve this, we (1) input the climate model data into the Deconfounding Bias Correction (BC) factor model to obtain the multi-cause hidden confounder which is essential for bias correction (we call this step ’Deconfounding’). Then (2) we use a hidden multicause confounder as a bias source, combine it with observational data, building a precipitation correction model to help the climate model to have a better estimate of future output (we call this step ’Correction’)\footnote{Code is avaliable at github: \url{https://github.com/Wentao-Gao/A-Factor-Model-Approach-to-Climate-Model-Bias-Correction}}.

\subsection{Deconfounding}

To address the challenge of deconfounding time series data with time-varying latent confounders, \cite{bica2020time} proposed the Time Series Deconfounder (TSD), which extends the deconfounder methodology introduced by \cite{Wang2019} to the time series domain. The fundamental principle of the TSD is that it utilizes factor model to infer substitutes for hidden confounders as treatments.

The goal of this part is to generalize the factor model proposed by \cite{bica2020time} to a Deconfounding BC factor model for bias correction. This extension aims to extend the Time Series Deconfounder to a complex time series setting of two sources.

\begin{figure}[h]
        \centering
        \includegraphics[width=0.8\linewidth]{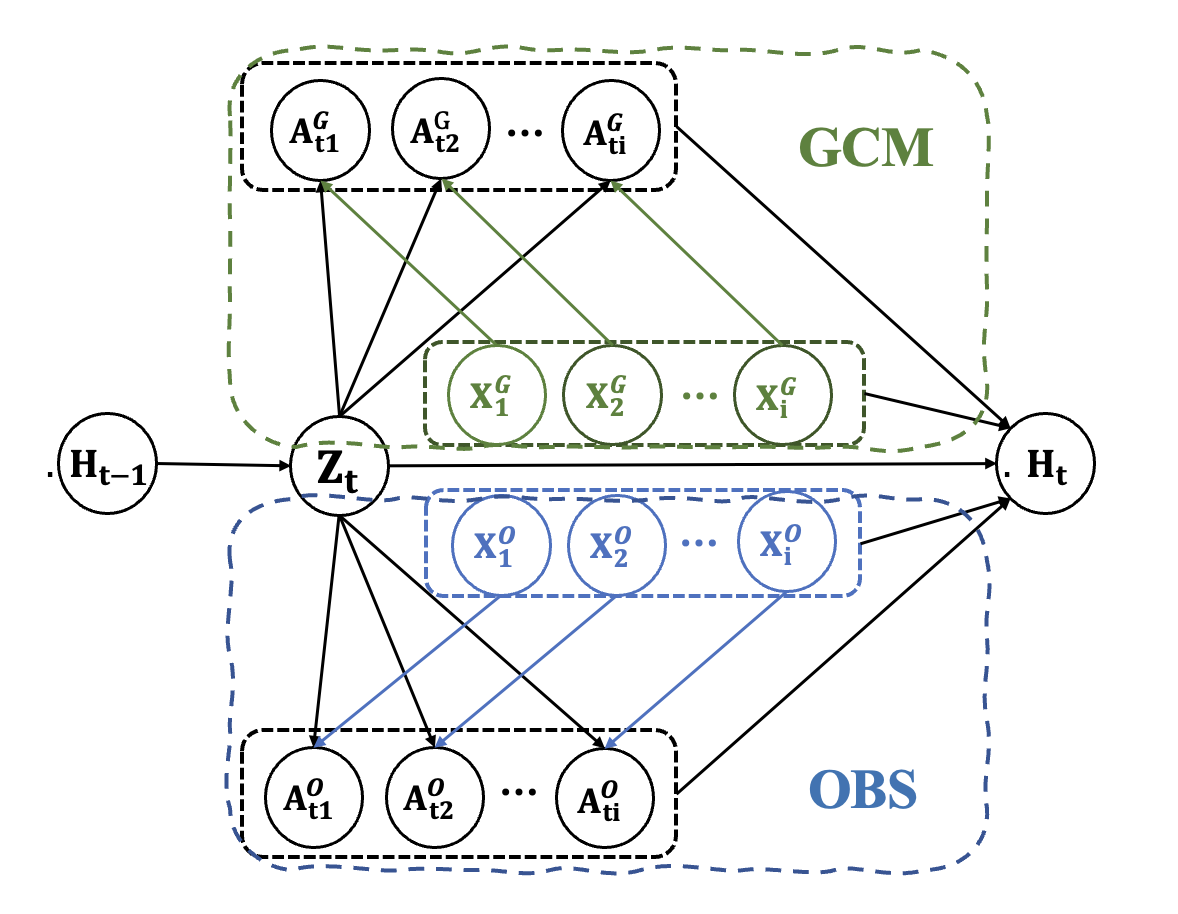}
        \caption{Causal graphs of two sources with latent confounders.}
        \label{fig:double source}
\end{figure}

\subsubsection{Deconfounding BC Factor Model}

To extend time series deconfounder~\cite{bica2020time} to bias correction, We propose an enhanced multi-source factor model specially for this task, termed the Deconfounding BC factor model. 

For single source data, the unobserved confounder affects \(\mathbf{X, A}, Y\) from one source, allowing us to infer the sequence of unobserved confounders \(\mathbf{z}_t = g(\bar{\mathbf{h}}_{t-1})\), where \(\bar{\mathbf{h}}_{t-1} = \{\bar{\mathbf{a}}_{t-1}, \bar{\mathbf{x}}_{t-1}, \bar{\mathbf{z}}_{t-1}\}\) is the realization of \(\bar{\mathbf{H}}_{t-1}\). Specifically, factorization can be expressed as follows:
\begin{equation}
    p(a_{t1}, \ldots, a_{tk} \mid \mathbf{z}_t, \mathbf{x}_t) = \prod_{j=1}^k p(a_{tj} \mid \mathbf{z}_t, \mathbf{x}_t).
\end{equation}

\begin{figure*}[h]
    \centering
    \includegraphics[width=1\linewidth]{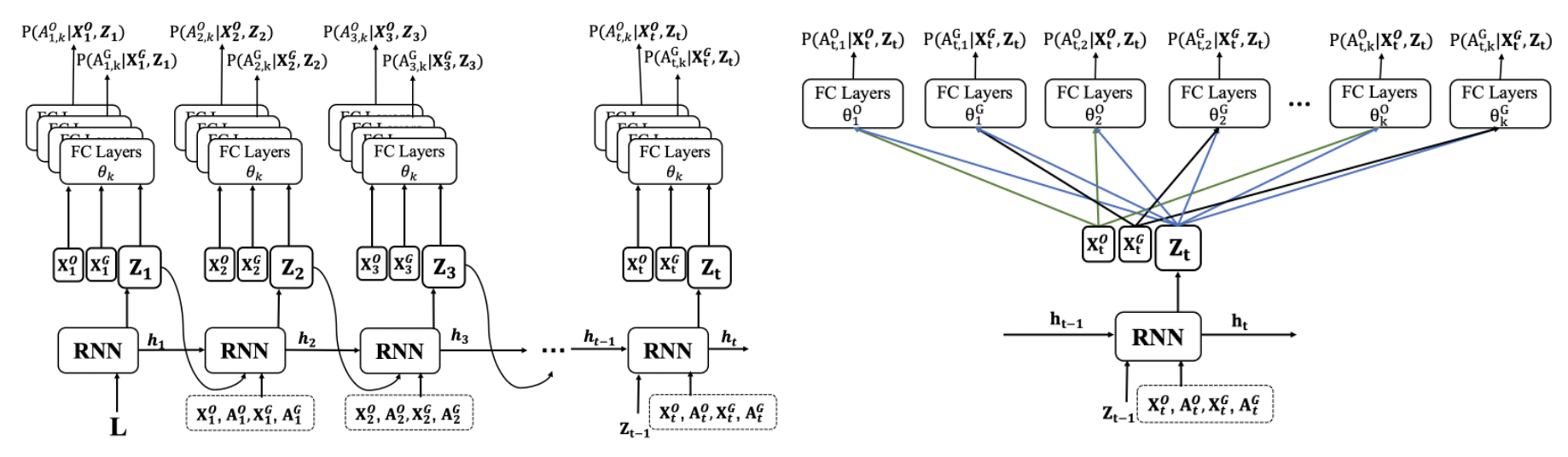}
    \caption{
    The architecture of the proposed Deconfounding BC factor model. 
    At each time step \( t \), a shared RNN takes the past treatment and covariate sequences from both GCM and observational sources as input and generates a latent confounder representation \( \mathbf{Z}_t \). 
    This latent confounder, along with the current covariates from GCM (\( \mathbf{X}_t^G \)) and observations (\( \mathbf{X}_t^O \)), is then fed into multiple fully connected layers (one per treatment variable) to predict the treatment distributions \( p(A_{tj}^G \mid \mathbf{X}_t^G, \mathbf{Z}_t) \) and \( p(A_{tj}^O \mid \mathbf{X}_t^O, \mathbf{Z}_t) \).
    By structuring the model, the learned \( \mathbf{Z}_t \) serves to deconfound both sources simultaneously, enabling conditionally independent treatment modeling given \( \mathbf{Z}_t \) and observed covariates.
    }
    \label{fig:factor-model}
\end{figure*}

To extend this model for multi-source data, consider both GCM and Observations as in Figure \ref{fig:double source}. The unobserved confounder will affect \(\mathbf{X, A, Y}\) in both sources, allowing us to infer the sequence of unobserved confounders \(\mathbf{z}_t = g(\bar{\mathbf{h}}_{t-1})\) that can be used to render both source treatments conditionally independent, where \(\bar{\mathbf{h}}_{t-1} = \{{\bar{\mathbf{h}}_{t-1}^O, \bar{\mathbf{h}}_{t-1}^G}\}\) is the realization of \(\bar{\mathbf{H}}_{t-1}\). Specifically, the factorization for the multi-source scenario can be expressed as follows:
\begin{equation}
p(a_{t1}^O,\dots,a_{tk}^O \mid \mathbf{z}_t,\mathbf{x}_t^O)
= \prod_{j=1}^k p(a_{tj}^O \mid \mathbf{z}_t,\mathbf{x}_t^O).
\end{equation}
\begin{equation}
p(a_{t1}^G, \ldots, a_{tk}^G \mid \mathbf{z}_t, \mathbf{x}_t^G) = \prod_{j=1}^k p(a_{tj}^G \mid \mathbf{z}_t, \mathbf{x}_t^G).
\end{equation}
It allows us to infer the sequence of latent variables \( \mathbf{Z}_t \) such that, conditioned on \( \mathbf{Z}_t \) and the covariates from both sources \( \mathbf{X}_t^G \) and \( \mathbf{X}_t^O \), the treatments become conditionally independent. As in one location, the same confounders \(\mathbf{Z}\) will affect both GCM and observations. In this case, the \(\mathbf{Z}\) we learned should render treatments in both GCM and observational data conditionally independent with covariate \(\mathbf{X_t}\). 

The structure of the factor model depends on causality, which relies on the assumptions listed below. 

\begin{assumption}
\textbf{Consistency.} If $\bar{A}_{\geq t} = \bar{a}_{\geq t}$, then the potential outcomes for following the treatment $\bar{a}_{\geq t}$ is the same as the factual outcome $Y(\bar{a}_{\geq t}) = Y$.
\end{assumption}

\begin{assumption}
\textbf{Positivity (Overlap)\cite{bica2020time}:} if \( P(\mathbf{\bar{A}}_{t-1} = \mathbf{\bar{a}}_{t-1}, \mathbf{X}_t = \mathbf{\bar{x}}_t) \neq 0 \) then \( P(A_t = a_t \mid \mathbf{\bar{A}}_{t-1} = \mathbf{\bar{a}}_{t-1}, \mathbf{X}_t = \mathbf{\bar{x}}_t) > 0 \) for all \( a_t \).
\end{assumption}

\begin{assumption}
\textbf{Sequential Single Strong Ignorability}
\begin{equation}
\mathbf{Y}(\bar{\mathbf{a}}_{\geq t}) \perp \!\!\! \perp {A}_{tj} \mid \mathbf{X}_t, \bar{\mathbf{H}}_{t-1}
\end{equation}
for all \( \bar{\mathbf{a}}_{\geq t} \), for all \( t \in \{0, \ldots, T\} \), and for all  j $\in$ \{1, \ldots, k\}.
\end{assumption}

Next, we provide a theoretical analysis for the soundness of the learned $\mathbf{Z}_t$ by introducing the concept of sequential Kallenberg construction~\cite{bica2020time}, as follows.

\begin{Definition}\textbf{Sequential Kallenberg construction}
At timestep \(t\), we say that the distribution of assigned causes \(({A}^G_{t1}, \ldots, {A}^G_{tk}), ({A}^O_{t1}, \ldots, {A}^O_{tk})\) admits a sequential Kallenberg construction from the random variables $\mathbf{Z}_t = g(\mathbf{\bar{H}}^O_{t-1}, \mathbf{\bar{H}}^G_{t-1})$ and $\mathbf{X}^O_t, \mathbf{X}^G_t$ if there exist measurable functions \(f_{tj}^O : \mathbf{Z^O_t} \times \mathbf{X^O_t} \times [0, 1] \to A_j^O\) and \(f_{tj}^G : \mathbf{Z^G_t} \times \mathbf{X^G_t} \times [0, 1] \to A_j^G\) and random variables \(U_{tj}^O, U_{tj}^G \in [0,1]\), with \(j = 1, \ldots, k\) such that:
\begin{equation}
    A_{tj}^O = f_{tj}^O(\mathbf{Z_t}, \mathbf{X^O_t}, U_{tj}^O)
\end{equation}
\begin{equation}
A_{tj}^G = f_{tj}^G(\mathbf{Z_t}, \mathbf{X^G_t}, U_{tj}^G),
\end{equation}
where \(U_{tj}^O, U_{tj}^G\) marginally follow \(\text{Uniform}[0, 1]\) and jointly satisfy:
\begin{equation}
\bigl(U_{t1}^O,\dots,U_{tk}^O\bigr)
\;\indep\;
Y^O\bigl(\bar{\mathbf a}^O_{\ge t}\bigr)
\;\bigm|\;
\mathbf Z_t,\;\mathbf X_t^O,\;\bar{\mathbf H}_{t-1}^O
\end{equation}
\begin{equation}
\bigl(U_{t1}^G,\dots,U_{tk}^G\bigr)
\;\indep\;
Y^G\bigl(\bar{\mathbf a}^G_{\ge t}\bigr)
\;\bigm|\;
\mathbf Z_t,\;\mathbf X_t^G,\;\bar{\mathbf H}_{t-1}^G
\end{equation}

for all \(\mathbf{\bar{a}}^O_{\ge t}\) and \(\mathbf{\bar{a}}^G_{\ge t}\).
\end{Definition}

We present the following theorem to guarantee that the learned $\mathbf{Z}_t$ can serve as a substitute for the multi cause hidden confounders. 
\begin{theorem}\textbf{The soundness of the learned $\mathbf{Z}_t$}
If at every timestep $t$, the two distribution of assigned causes $(\mathbf{A}^G_{t1}, \ldots, \mathbf{A}^G_{tk})$, $(\mathbf{A}^O_{t1}, \ldots, \mathbf{A}^O_{tk})$  admit a Kallenberg construction from $\mathbf{Z}_t = g(\mathbf{\bar{H}}^O_{t-1}, \mathbf{\bar{H}}^G_{t-1})$ and $\mathbf{X}^O_t, \mathbf{X}^G_t$ separately, then the learned $\mathbf{Z}_t$ can serve as a substitute for the multi cause hidden confounders. 
\end{theorem}

Theorem~1 establishes that the latent variable \( \mathbf{Z}_t \) can be inferred by leveraging the conditional independence structure of the treatments: within each source (GCM or Observation), the assigned treatments \( \{A_{t1}, \ldots, A_{tk}\} \) are conditionally independent given \( \mathbf{Z}_t \) and their respective covariates \( \mathbf{X}_t^G \) and \( \mathbf{X}_t^O \). This structure, illustrated in Figure~\ref{fig:factor-model}, provides the necessary identifiability condition for inferring \( \mathbf{Z}_t \) through a shared factor model across both sources. The result means that, at each timestep, for both sources, the variable \noindent $\mathbf{\bar{X}}_t$, $\mathbf{\bar{Z}}_t$, $\mathbf{\bar{A}}_{t-1}$ contain all of the dependencies between the potential outcomes $\mathbf{Y}(\mathbf{\bar{a}}_{\geq t})$ and the assigned causes $\mathbf{A}_t$.

Then for both sources we have,
\begin{equation}
    {E}[Y(\bar{a}_{\geq t}) \mid \bar{\mathbf{A}}_{t-1}, \bar{\mathbf{X}}_t, \bar{\mathbf{Z}}_t] = {E}[Y \mid \bar{a}_{\geq t}, \bar{\mathbf{A}}_{t-1}, \bar{\mathbf{X}}_t, \bar{\mathbf{Z}}_t].
\end{equation}
This equation implies that the potential outcome can be represented by
\({E}[Y \mid \bar{a}_{\geq t}, \bar{\mathbf{A}}_{t-1}, \bar{\mathbf{X}}_t, \bar{\mathbf{Z}}_t]\). Therefore, our estimate of the potential outcome is unbiased given the previous treatments, current covariates, and latent variables. This unbiasedness is crucial for making valid inferences about causal effects.

Then, we will show how to implement in practical based on factorization Formula (3) and (4). The joint distribution of all collected T time steps in our enhanced Deconfounding BC Factor Model can be formulated as follows:
\begin{equation}
\begin{aligned}
p(\theta_{1:k}^G, \bar{\mathbf{x}}_T^G, \bar{\mathbf{z}}_T, \bar{\mathbf{a}}_T^G) 
= & \, p(\theta_{1:k}^G)p(\bar{\mathbf{x}}_T^G) 
\prod_{t=1}^T p(\mathbf{z}_t \mid \bar{\mathbf{h}}_{t-1}) \\
& \times \prod_{j=1}^k p(\mathbf{a}_{tj}^G \mid \mathbf{z}_t, \mathbf{x}_t^G, \theta_j^G),
\end{aligned}
\end{equation}
The same formulation holds for the GCM source by replacing \(O\) with \(G\).

\begin{figure*}[h]
    \centering
    \includegraphics[width=1.0\linewidth]{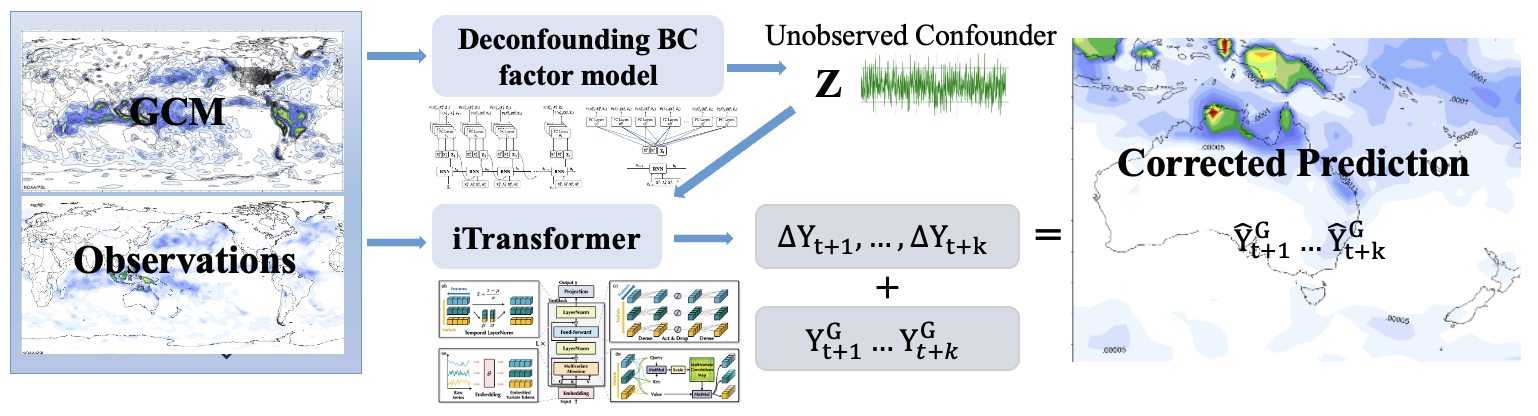}
    \caption{Workflow of the Deconfounding BC factor model(model architecture refers to Figure \ref{fig:factor-model}) using simulation data: the GCM Outputs and observational data are combined to estimate the unobserved confounder \( Z \) through the Deconfounding BC factor model. The iTransformer utilizes \( Z \) to compute corrections (\( \Delta Y_{t+1}, \ldots, \Delta Y_{t+k} \)) and adds them to the GCM predictions (\( Y_{t+1}^G, \ldots, Y_{t+k}^G \)) to generate the final corrected precipitation predictions.}
    \label{fig:overallprocess}
\end{figure*}

where \( \theta_{1:k}^O \) and \( \theta_{1:k}^G \) are the model parameters. The treatment distributions \( p(\mathbf{\bar{a}}_T^O) \) and \( p(\mathbf{\bar{a}}_T^G) \) are the corresponding marginal distributions.

In practical applications, as shown Figure \ref{fig:factor-model}, we need to build a model to obtain \(\mathbf{Z}\) with predicting treatments as the constraint. To manage time-varying treatments \(\mathbf{A}^G, \mathbf{A}^O\) , we follow a pragmatic approach by utilizing a recurrent neural network (RNN) with multitask output for implementation.

The recurrent model infers the latent variables \( \mathbf{Z_t} \) based on location history:
\begin{equation}
\mathbf{Z}_1 = \text{RNN}(\mathbf{L}), 
\end{equation}
\begin{equation}
\quad \mathbf{Z}_t = \text{RNN}(\mathbf{Z}_{t-1}, \mathbf{X}_{t-1}^G, \mathbf{A}_{t-1}^G, \mathbf{X}_{t-1}^O, \mathbf{A}_{t-1}^O, \mathbf{L}),
\end{equation}
where \( L \) consists of randomly initialized trainable parameters. Note that
in each time step \( t \), the assignments \( \mathbf{A}_t^O = [A_{t1}^O, \dots, A_{tk}^O] \) and \( \mathbf{A}_t^G = [A_{t1}^G, \dots, A_{tk}^G] \) are conditionally independent given \( \mathbf{Z_t}\), \( \mathbf{X_t}^O \) and \( \mathbf{Z_t}\), \( \mathbf{X_t}^G \) separately:
\begin{equation}
    A_{tj}^G = \text{FC}(\mathbf{X}_t^G,\mathbf{Z_t},\theta_j^G),
\end{equation}
\begin{equation}
    A_{tj}^O = \text{FC}(\mathbf{X}_t^O, \mathbf{Z}_t, \theta_j^O),
\end{equation}
for all \( j \) and \( t \), where \( \theta_j^G \) and \( \theta_j^O \) are the parameters of the fully connected layers. Using these as multitask output, treatments is conditionally independent given \(\mathbf{X_t, Z_t}\) for both sources, this can be ensured by inferring each treatment $A^G$ or $A^O$ as different tasks with only $\mathbf{X}_t^G,\mathbf{Z_t}$ or $\mathbf{X}_t^O,\mathbf{Z_t}$, if we condition on $\mathbf{X}_t^G,\mathbf{Z_t}$ or $\mathbf{X}_t^O,\mathbf{Z_t}$, they are conditionally independent.

\subsubsection{Training} 
The factor model is trained using gradient descent methods on the observational and GCM dataset. The architecture leverages the dependencies between multiple treatments and the location history to infer the latent variables.

\subsection{Correction}

After obtaining hidden multi-cause confounders \(\mathbf{Z}\) from both GCM and observational data in one area. In our problem formulation, we aim to correct the bias between \(Y^G\) and \(Y^O\) while considering unobserved confounders. A key question is how to utilize this information for bias correction. Advanced time series forecasting models such as iTransformer \cite{liu2024itransformerinvertedtransformerseffective} have demonstrated significant potential in capturing temporal information, and integrating these models into our bias correction approach promises substantial improvements.  Compared to models like TimeMixer and PatchTST, iTransformer balances complexity and performance.

We propose to use a probability model for bias correction. As shown in Figure \ref{fig:overallprocess}, we build a prediction model based on the observational data and the multi-cause confounder Z to predict future precipitation.

The probability model employed in our framework is designed to generate predictions of \( \Delta Y \), the correction term for the GCM output. Formally, the model learns to approximate the conditional distribution:
\(
P_\theta(\Delta Y_{t+k} \mid \mathbf{A}^G, \mathbf{A}^O, \mathbf{Z}),
\)
where \( \mathbf{A}^G \) and \( \mathbf{A}^O \) are the current covariates from GCM and observations, and \( \mathbf{Z} \) is the inferred latent confounder.

To achieve high performance in forecasting tasks involving complex temporal dependencies, we choose to use iTransformer \cite{liu2024itransformerinvertedtransformerseffective} as the prediction model. iTransformer is a lightweight and effective transformer variant specifically designed for long-sequence time series forecasting, offering a strong balance between expressiveness and efficiency.

After we obtain the \(\Delta{Y}_{t+k}\), we can derive our corrected GCM output with:
\begin{equation}
\begin{split}
\{\hat{Y}_{t+1}^{G}, \hat{Y}_{t+2}^{G}, \dots, \hat{Y}_{t+k}^{G}\} = \\
 \{Y_{t+1}^{G} + \Delta Y_{t+1}, 
Y_{t+2}^{G} + \Delta Y_{t+2}, \dots, Y_{t+k}^{G} + \Delta Y_{t+k}\}
\end{split}
\end{equation}

\section{Experiments And Results}

The objectives of the experiments are as follows: 
1) Use synthetic data sets to evaluate the correctness of the latent confounder captured by our deconfounding method.
2) Build a correction model to correct precipitation predictions made by the climate model. Implementation code is attached.

\subsection{Experiments On Synthetic Data}

To assess the effectiveness of our deconfounding method, we conduct controlled experiments on synthetic data, since the true influence of hidden confounders cannot be measured directly in real‐world settings \cite{Wang2019}. We generate data from a two‐source autoregressive model based on the causal summary graph in Figure 1(c). At each timestep \(t\), we sample time‐varying covariates \(X_t^{\text{source1}}\) and \(X_t^{\text{source2}}\), together with a multi‐cause latent confounder \(Z_t\). We simulate \(N=500\) locations over \(T=3650\) timesteps, with \(k=3\) covariates and treatments and autoregressive order \(p=5\). The resulting dataset is split into 80\% training, 10\% validation, and 10\% testing.

\subsubsection{Simulated Dataset}

\begin{figure*}[h]
    \centering
    \begin{subfigure}[b]{0.33\linewidth}
        \centering
        \includegraphics[width=\linewidth]{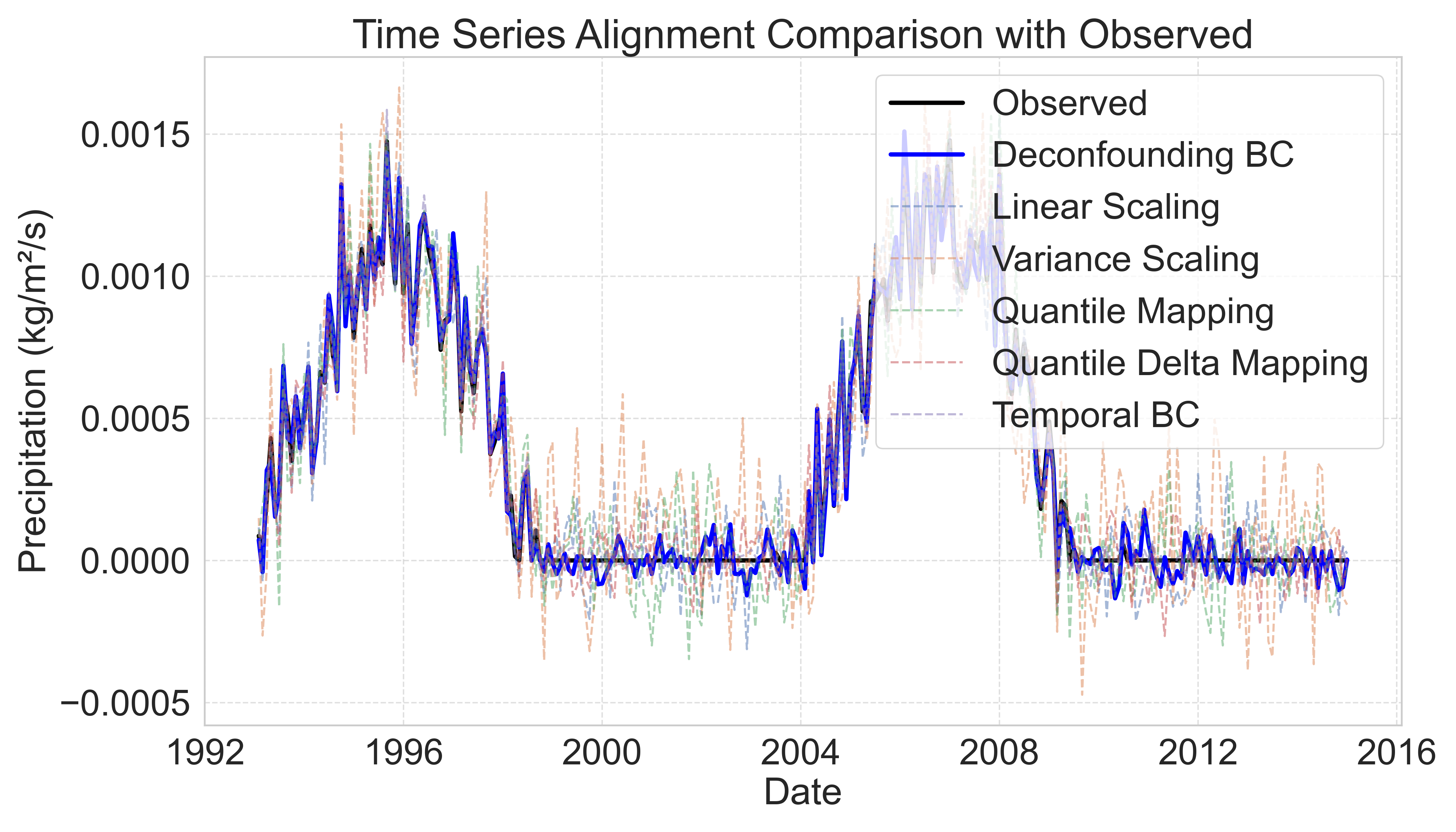}  
        \caption{Time series plot}
        \label{fig:qq_plot_bc1}
    \end{subfigure}
    \hfill 
    \begin{subfigure}[b]{0.33\linewidth}
        \centering
        \includegraphics[width=\linewidth]{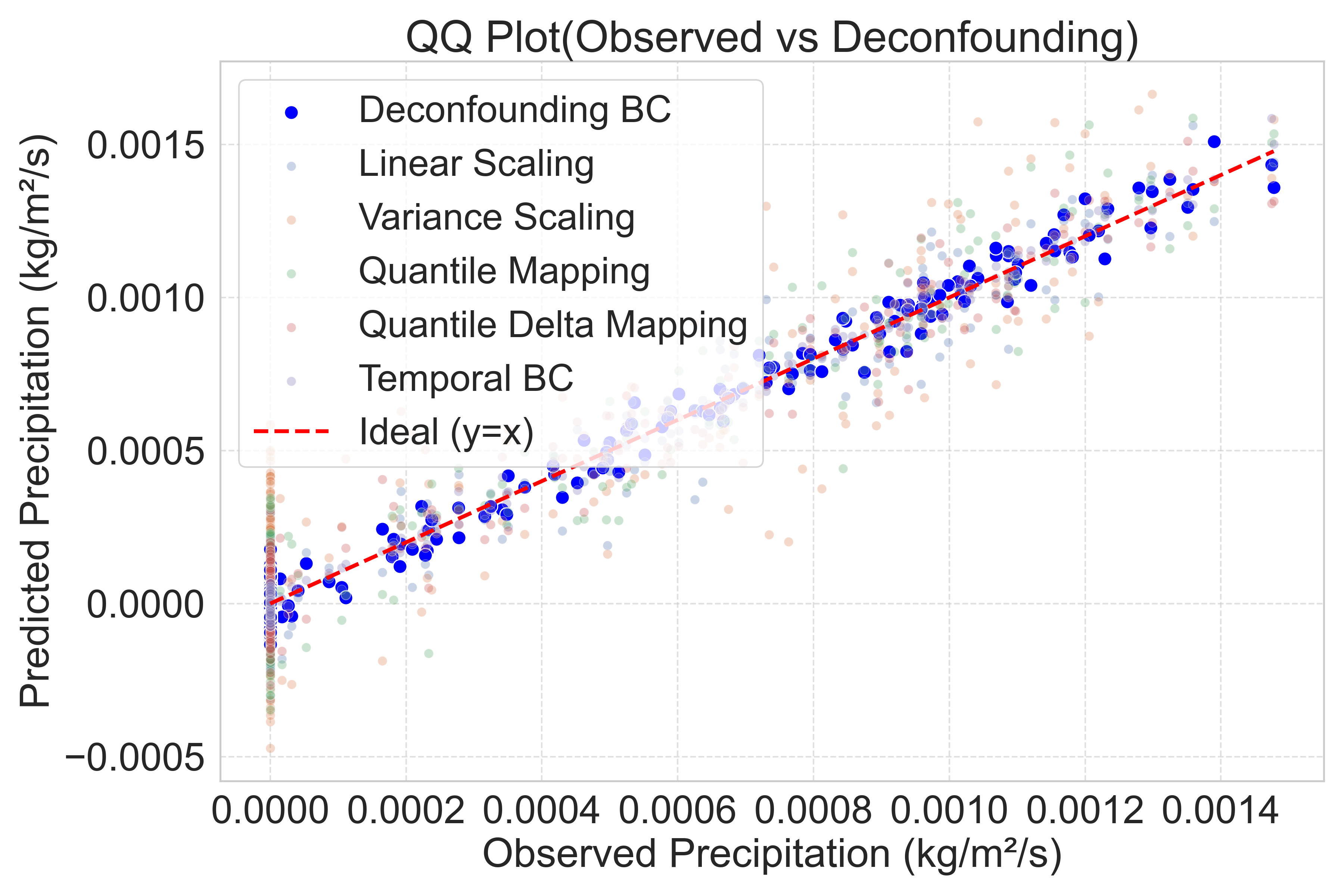}  
        \caption{QQ plot}
        \label{fig:qq_plot_bc2}
    \end{subfigure}
    \hfill
    \begin{subfigure}[b]{0.33\linewidth}
        \centering
        \includegraphics[width=\linewidth]{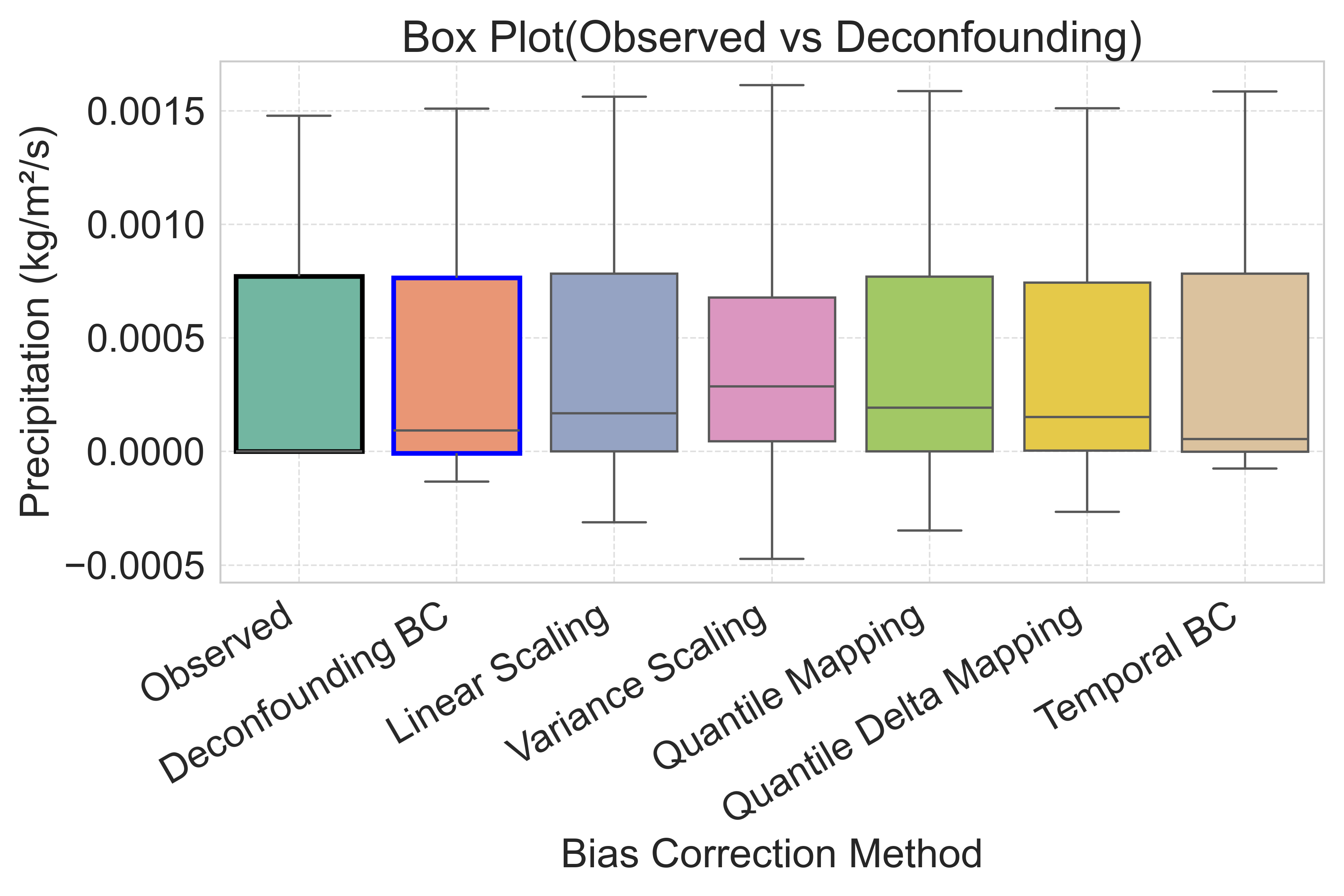}  
        \caption{Box plot}
        \label{fig:qq_plot_bc3}
    \end{subfigure}
    
    \caption{Comparison of bias correction methods for precipitation using simulation data: (Left) Time series alignment with observed data, demonstrating the performance of Deconfounding BC and other methods; (Center) QQ plot comparing predicted rainfall against observed values, highlighting Deconfounding BC's closer alignment with the ideal line; (Right) Box plot of rainfall distributions across bias correction methods, illustrating variability and alignment with observed data.}
    \label{fig:all}
\end{figure*}

Our evaluation focuses on two properties. First, the inferred latent variable \(\mathbf{Z}\), together with the observed covariates \(\mathbf{X}\), should accurately predict the treatment vector \(\mathbf{A}=[A_1,\ldots,A_k]\). Concretely, we model
\[p(\mathbf{A}\mid \mathbf{Z},\mathbf{X})
=\prod_{j=1}^k p\bigl(A_j \mid \mathbf{Z},\mathbf{X}\bigr)\].

Second, the learned \(\mathbf{Z}\) should match the ground‐truth confounders used in simulation. A low mean squared error (MSE) between inferred and true \(\mathbf{Z}\) indicates successful recovery of the hidden structure.
The details for data generation can be found in \href{https://github.com/Wentao-Gao/A-Factor-Model-Approach-to-Climate-Model-Bias-Correction}{our GitHub repository}.

Figure~\ref{fig:all} summarizes our predictive checks. The Deconfounding BC method yields an MSE of \(0.06157\) when predicting the treatment vector, demonstrating strong alignment between model predictions and the simulated treatments. Comparing the inferred \(\mathbf{Z}\) against the true latent confounders gives an MSE of \(0.00187\), confirming that our factor‐model approach accurately captures the hidden variables. Together, these results validate that the treatments become conditionally independent given the inferred confounders and covariates, and that the learned latent representation faithfully recovers the ground‐truth confounding signal.

\subsection{Case Study: South Australia}

In a bias correction (BC) process, data from climate models are compared to actual observations (or their proxies, like reanalysis products) to adjust for biases. Since global climate models (GCMs) usually have a lower resolution compared to observations or reanalysis reference data, BC often involves downscaling the resolution of GCMs. We adopt this methodology as well. For the climate model data, we selected the 15 initial condition runs from the Institut Pierre-Simon Laplace (IPSL) climate model as part of the sixth Coupled Model Intercomparison Project (CMIP6) historical experiment. The data, which is available on a monthly basis, includes climate variables such as tmax (maximum temperature 2 meters above the surface) and prate (precipitation rate), with our bias correction efforts centered on prate.

The IPSL\footnote{IPSL data portal: https://aims2.llnl.gov/search/cmip6/} model is run at a 250km nominal resolution and is not re-gridded. We selected the closest geographical point to South Australia for the case study, covering the period from 1948 to 2014.

\begin{table*}[htbp]
    \centering
    \begin{tabular}{lcccc}
        \toprule
        \textbf{Method} & \textbf{Exp 1} & \textbf{Exp 2} & \textbf{Exp 3} & \textbf{Average} \\
        & \textbf{MSE / MAE} & \textbf{MSE / MAE} & \textbf{MSE / MAE} & \textbf{MSE / MAE} \\
        \midrule
        IPSL & 0.076 / 0.217 & 0.0507 / 0.1731 & 0.0191 / 0.1059 & 0.0486 / 0.1653 \\
        Linear Scaling & 0.0373 / 0.148 & 0.0423 / 0.1587 & 0.0379 / 0.1471 & 0.0392 / 0.1512 \\
        Variance Scaling & 0.0055 / 0.0541 & 0.0072 / 0.0614 & 0.0113 / 0.0782 & 0.0080 / 0.0646 \\
        Quantile Mapping & 0.0479 / 0.166 & 0.0413 / 0.1535 & 0.0473 / 0.1637 & 0.0455 / 0.1611 \\
        Quantile Delta Mapping & 0.0347 / 0.141 & 0.0320 / 0.1348 & 0.0140 / 0.0885 & 0.0270 / 0.1211 \\
        Temporal BC & 0.0042 / 0.0412 & 0.0047 / 0.0327 & 0.0049 / 0.0411 & 0.0046 / 0.0383 \\
        Deconfounding BC & \textbf{0.0018} / \textbf{0.0175} & \textbf{0.0016} / \textbf{0.01615} & \textbf{0.0025} / \textbf{0.01529} & \textbf{0.0020} / \textbf{0.01632} \\
        \bottomrule
    \end{tabular}
    \caption{Comparison of Bias Correction Methods Across Multiple Experimental Conditions. Exp 1, Exp 2, and Exp 3 correspond to the climate model outputs based on r5i1p1f1, r6i1p1f1, and r7i1p1f1, respectively. For Temporal BC and Deconfounding BC, a 36-month history input sequence length and a 3-month future output sequence length were chosen. Bolded values represent the best results in each experiment.}
    \label{tab:comparison}
\end{table*}

\begin{table*}[h]
    \centering
    \begin{tabular}{lcccccccc}
        \toprule
        & \textbf{Obs} & \multicolumn{2}{c}{\textbf{Exp1 (r5i1p1f1)}} & \multicolumn{2}{c}{\textbf{Exp2 (r6i1p1f1)}} & \multicolumn{2}{c}{\textbf{Exp3 (r7i1p1f1)}} \\
        \textbf{Metric} & & \textbf{With Z} & \textbf{Without Z} & \textbf{With Z} & \textbf{Without Z} & \textbf{With Z} & \textbf{Without Z} \\
        \midrule
        MSE & 0.00848 & \textbf{0.001771} & 0.004275 & \textbf{0.001632} & 0.002162 & \textbf{0.002523} & 0.002995 \\
        MAE & 0.0186 & \textbf{0.01751} & 0.03346 & \textbf{0.01615} & 0.01840 & \textbf{0.01529} & 0.02323 \\
        \bottomrule
    \end{tabular}
    \caption{Comparison of Correction Model Results With and Without Latent Confounder Across Multiple Experimental Conditions. Bolded values represent the best results in each experiment.}
    \label{tab:comparison_latent_confounder}
\end{table*}

For observational reference data, NCEP-NCAR Reanalysis 1\footnote{NCEP-NCAR data portal: https://psl.noaa.gov/data/gridded/data.
ncep.reanalysis.html}, provided by the National Oceanic and Atmospheric Administration (NOAA), is utilized. The dataset encompasses the same variables as the IPSL model (e.g., tmax \& prate). NCEP-NCAR reanalysis data is available at a monthly frequency, covering the period from 1948 to 2014, with a global resolution of 2.5 degrees in both the latitudinal and longitudinal directions, and has not been re-gridded. The nearest geographical point to South Australia was selected.

This research focuses on bias correction of the precipitation rate to improve the accuracy of precipitation predictions.

\subsubsection{Data Preprocessing}
We extracted and converted climate data from 1948 to 2014 for the South Australia region. The data, originally in NetCDF format, were transformed into CSV files for further processing. Subsequently, we split the data into a training period (1948-1992) and a testing period (1993-2014). The IPSL dataset contains 30 initial conditions; to generalize our model, we selected 15 initial conditions for the experiment. In this paper, we present results from three of these initial conditions(r5i1p1f1, r6i1p1f1 and r7i1p1f1) along with their average results. 

\subsubsection{Experiment Settings}To evaluate our proposed method, we compare our method with several bias correction baseline methods, including linear scaling \cite{Teutschbein2012}, variance scaling \cite{Teutschbein2012}, quantile mapping \cite{cannon2015bias}, quantile delta mapping \cite{tong2021bias} and Temporal BC \cite{nivron2024temporalstochasticbiascorrection} methods. The metric we are using to compare the performance is the Mean Squared Error (MSE) and the Mean Absolute Error (MAE). 

\subsection{Results}

\begin{figure}[h]
    \centering
    \includegraphics[width=1.0\linewidth]{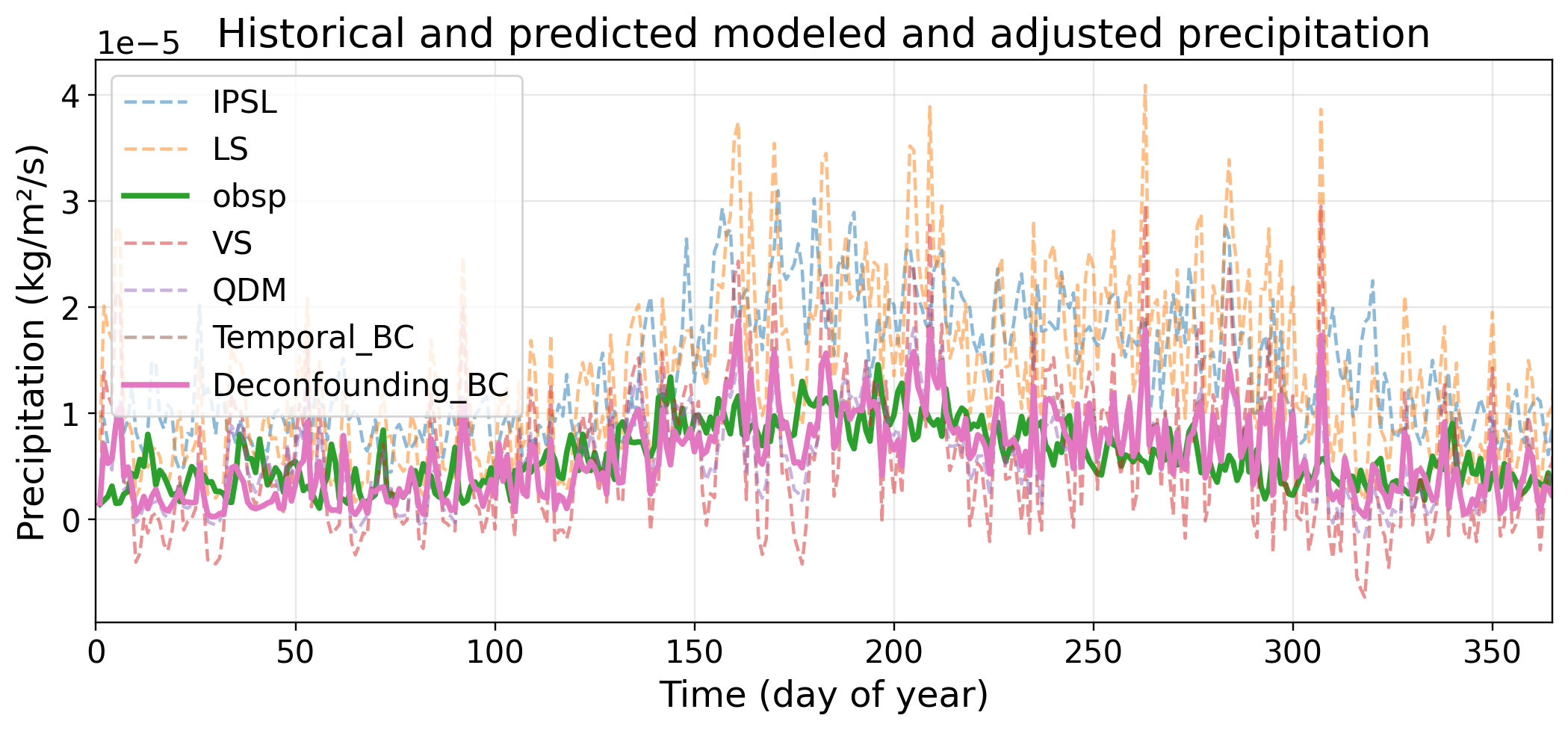}  
    \caption{Comparison of daily precipitation rates from IPSL , observation and various bias correction methods over a year, highlighting the performance of Deconfounding BC in aligning with observed trends.}
    \label{fig:figure6_precipitation}
\end{figure}

Table \ref{tab:comparison} illustrates the three-step-ahead predictions using our deconfounding bias correction method on the dataset with different GCM initial settings. After applying the deconfounding step and augmenting the dataset with substitutes for the hidden confounders, we evaluated the performance of various bias correction methods using MSE and MAE. The results in Table \ref{tab:comparison} support the effectiveness of the Deconfounding BC method, which consistently achieves the lowest MSE and MAE across all experimental conditions. This indicates its superior performance in reducing prediction errors and aligning closely with the observed data. Figure~\ref{fig:figure6_precipitation} further visualizes the daily precipitation trends from IPSL, observations, and various correction methods, clearly highlighting the superior alignment achieved by Deconfounding BC.

\begin{figure}[h]
    \centering
    \includegraphics[width=1\linewidth]{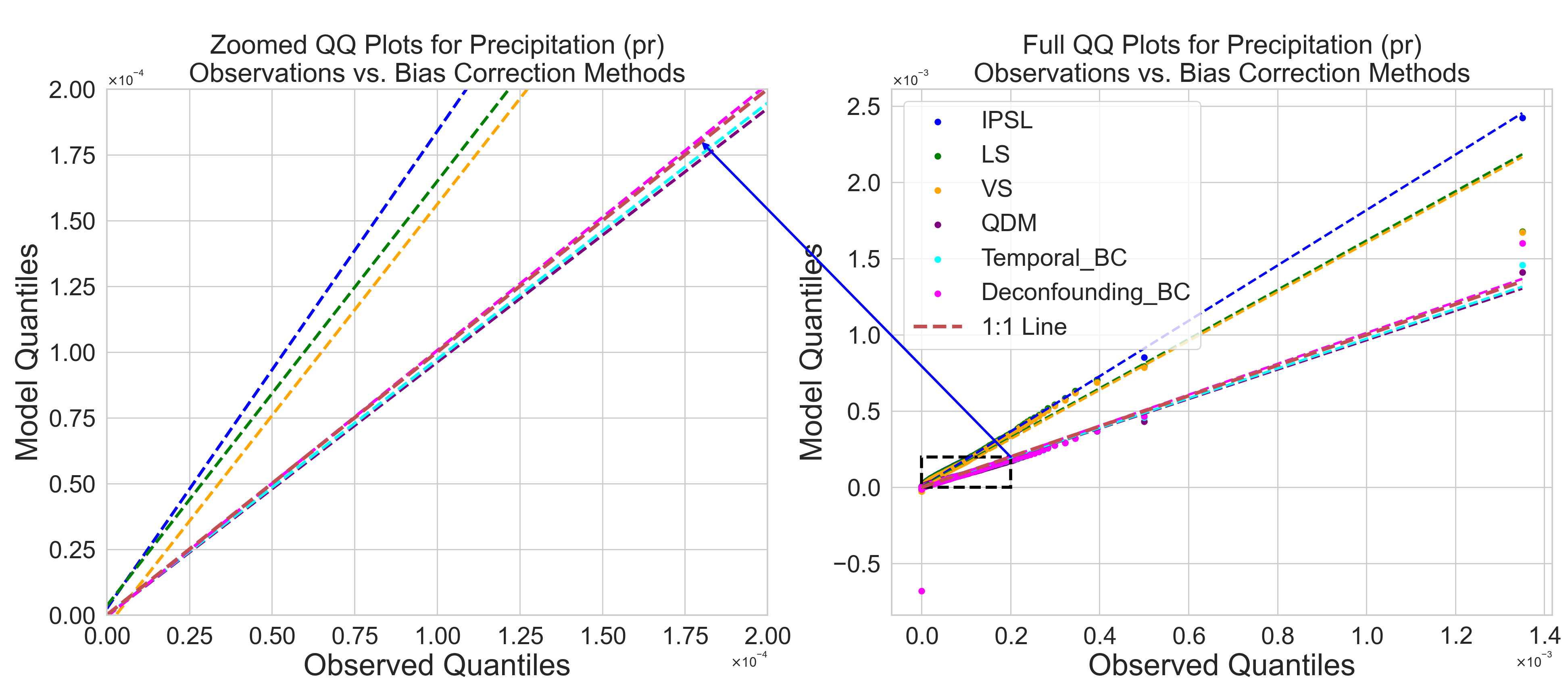}  
    \caption{This figure shows Quantile-Quantile (QQ) plots comparing observed and model precipitation quantiles using various bias correction methods. The left panel highlights the performance of different bias correction methods, with a focus on how closely they align with the 1:1 line, which indicates perfect agreement. The right panel illustrates overall agreement across all quantiles, providing a broader comparison of the methods.}
    \label{fig:qq_plots_bias_correction}
\end{figure}

The QQ plots (Figure \ref{fig:qq_plots_bias_correction}) reveal that among the evaluated bias correction methods, Deconfounding BC shows the closest alignment with the red dashed line (y=x) across the entire range of observed precipitation values. This indicates that Deconfounding BC is the best in accurately capturing the observed precipitation levels, outperforming other methods in reducing biases and providing reliable predictions.

To demonstrate that the latent confounder learned by our Deconfounding BC factor model contains essential information, we did ablation study to compare our correction model's results with and without the latent confounder, as shown in Table \ref{tab:comparison_latent_confounder}. The performance improved with the inclusion of the hidden confounder.

\section{Conclusion}

In this paper, we propose the deconfounding bias correction method for multi-cause confounders. By integrating climate bias correction techniques with causality-based time series deconfounding, our approach provides a novel perspective for future studies, emphasizing the importance of not assuming all variables are observed.

\section*{Acknowledgments}
This work was supported by the ARC Discovery Project DP230101122 and the University of South Australia Research Training Program (RTP) Scholarship. We gratefully acknowledge the continued support from the CSIRO Environment Research Unit and the Data61 Business Unit.

\bibliographystyle{named}
\bibliography{ijcai25}

\end{document}